\documentclass[letterpaper, 10 pt, conference]{ieeeconf}
\IEEEoverridecommandlockouts
\usepackage{times}
\usepackage{epsfig}
\usepackage{graphicx}
\usepackage{amsmath}
\usepackage{amssymb}
\usepackage{caption}
\usepackage{subcaption}
\usepackage{bm}
\usepackage{multirow}
\usepackage{color}
\usepackage{amsfonts}
\usepackage[utf8]{inputenc}

\usepackage[pagebackref=true,breaklinks=true,colorlinks,bookmarks=false]{hyperref}

\begin{document}

\captionsetup[figure]{name={Fig.},labelsep=period,singlelinecheck=off}

\title{\LARGE \bf
MotionHint: Self-Supervised Monocular Visual Odometry\\
with Motion Constraints}

\author{Cong Wang$^{1}$, Yu-Ping Wang$^{1}$, Dinesh Manocha$^{2}$
\thanks{$^{1}$Cong Wang and Yu-Ping Wang are with the Department of Computer Science and Technology, Tsinghua University, Beijing, China. }%
\thanks{$^{2}$Dinesh Manocha is with the Departments of Computer Science and Electrical \& Computer Engineering at the University of Maryland, MD 20742, USA. }%
\thanks{Yu-Ping Wang is the corresponding author, e-mail: wyp@tsinghua.edu.cn}%
\thanks{This work
was supported by the Natural Science Foundation of China (Project Number 61872210)}
\\
\url{https://github.com/JohnsonLC/MotionHint}
}

\maketitle

\begin{abstract}

We present a novel self-supervised algorithm named \emph{MotionHint} for monocular visual odometry (VO) that takes motion constraints into account.
A key aspect of our approach is to use an appropriate motion model that can help existing self-supervised monocular VO (SSM-VO) algorithms to overcome issues related to the local minima within their self-supervised loss functions.
The motion model is expressed with a neural network named \emph{PPnet}.
It is trained to coarsely predict the next pose of the camera and the uncertainty of this prediction.
Our self-supervised approach combines the original loss and the motion loss, which is the weighted difference between the prediction and the generated ego-motion.
Taking two existing SSM-VO systems as our baseline, we evaluate our \emph{MotionHint} algorithm on the standard KITTI benchmark.
Experimental results show that our \emph{MotionHint} algorithm can be easily applied to existing open-sourced state-of-the-art SSM-VO systems to greatly improve the performance by reducing the resulting ATE by up to 28.73\%.

\end{abstract}

\section{Introduction} \label{sec:introduction}

Visual Odometry (VO), is an essential task for many real-world applications such as autonomous driving, augmented reality, and robot navigation.
This problem has been extensively studied, and many geometric methods have been proposed~\cite{ORB22017, DSO2016, ORB2015, LSD2014, SVO2014, DBLP:conf/iros/WangZWDQM21, DBLP:conf/iccv/WangWM23, DBLP:conf/siggrapha/WangLBXZW19}.
However, geometric methods are based on rules of multi-view geometry and rely on hand-crafted features, which usually suffer from issues related to texture-less regions, blurred images, and ill-posed problems.
To overcome these limitations, recent research has exploited learning-based methods, which employ deep neural networks to improve the performance.
Supervised methods have resulted in promising results~\cite{firstCNN2015, DeepVO2017, Xue2018}. 
However, for supervised methods, the ground truth is required for training, and this is difficult to obtain in practice.


As an alternative, self-supervised methods have been proposed~\cite{bian2019, monodepth2, UnDeepVO2018, DeMoN2017, TartanVO2020, D3VO2020, zhou2017, DBLP:conf/siggrapha/WangKCBSZ23, DBLP:conf/cvpr/WangKSQWBZ25, DBLP:journals/corr/abs-2508-09597}.
Self-supervised methods can be trained using only monocular videos or synchronized stereo pairs, and they are able to predict the depth and the ego-motion together.
Self-supervision can be achieved using view synthesis and photometric error.
However, such self-supervision is likely to trap the VO system into local minima~\cite{DepthHint2019}.
We need extra information and constraints to avoid local minima.

{\bf Main Results:} We present a novel self-supervised method for monocular visual odometry. Our approach exploits the property that the trajectory of a camera must comply with the motion constraints of a vehicle (e.g., automobiles, ground robots, or drones) on which the camera is deployed.
A motion model that expresses such motion constraints could be used to improve the performance.
Based on this observation, there are two main issues: (1) How to define the motion model?; (2) How to apply the motion model to existing systems?
We propose a novel method, \emph{MotionHint}, to address both these issues and use it to improve the performance of self-supervised monocular VO (SSM-VO) systems.

\begin{figure}
\includegraphics[scale=0.31]{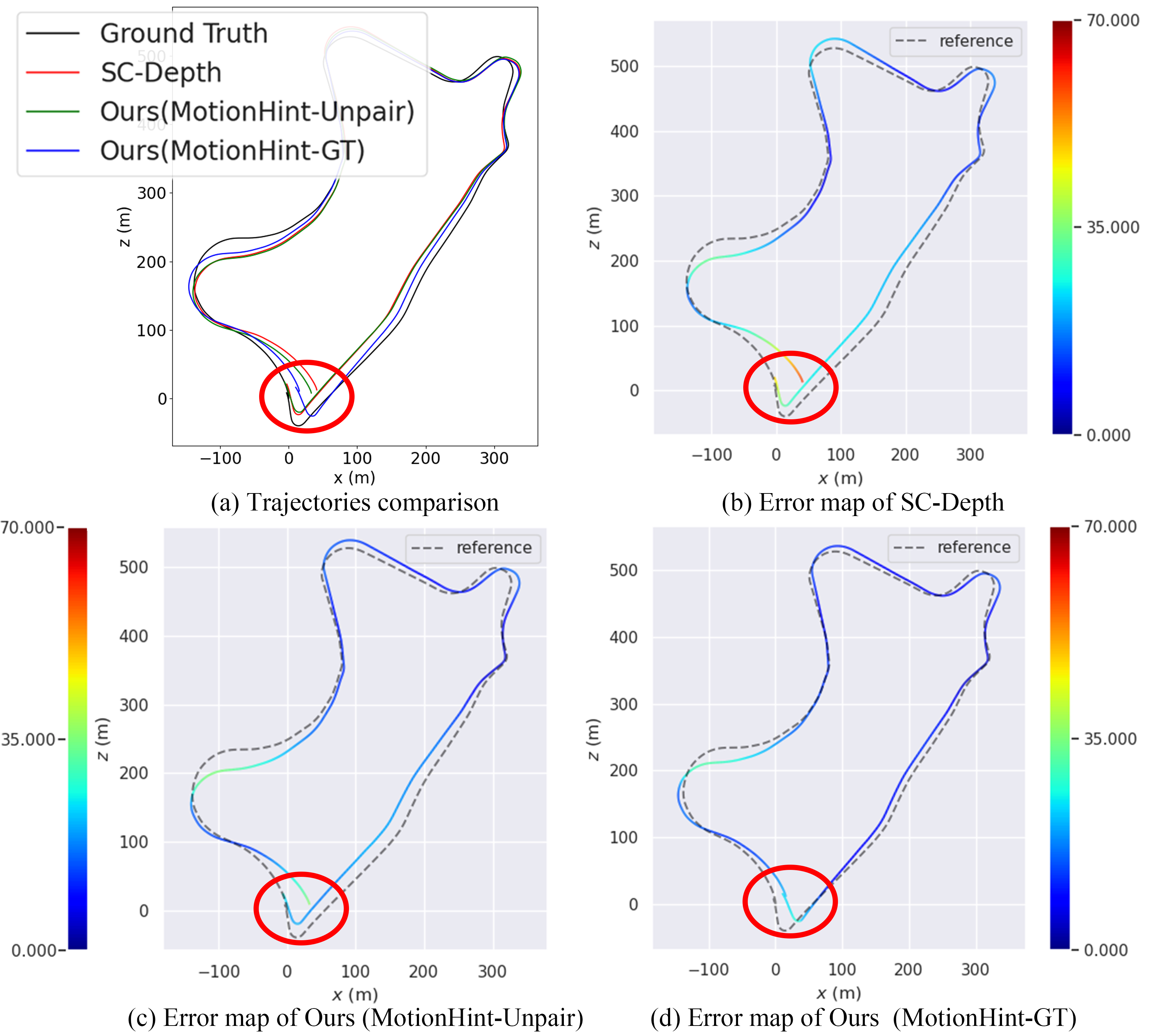}
   \caption{{\bf Trajectories Comparison.} 
   (a) demonstrates trajectories of SC-Depth~\cite{bian2021} (our baseline SSM-VO) and our improved version. 
   (b), (c) and (d) provide the error map of different predicted trajectories using the evo toolbox~\cite{evo2017}.
   Our \emph{MotionHint} algorithm can greatly improve the performance of SC-Depth by reducing resulting ATE by about 25\%, especially at the part in the red circles.
   } 
\label{fig:teaser}
\end{figure}

In order to address the first issue, we describe the motion model as a multivariate time series regression model (described in Section~\ref{sec:ppnet}).
We design \emph{PPnet} to express this motion model, which can generate the next pose and its uncertainty based on a set of consecutive prior poses.
\emph{PPnet} can be trained with any input pose sequences, either generated by geometric methods using any monocular videos captured by the same vehicle setup, or synthetized by simulations of the same vehicle setup.
In practice, such input pose sequences can be obtained more easily than the ground truth.

In order to deal with the second problem, we improve the self-supervision by considering the motion loss, which is the weighted difference between the ego-motion predicted by the original SSM-VO and the pseudo label generated by \emph{PPnet}.
During each training step, \emph{PPnet} takes consecutive poses predicted by the original SSM-VO as inputs and predicts the next pose to generate the pseudo label of current predicted ego-motion.
Finally, to combine the original loss and the motion loss, we adopt the form of a weighted sum and employ the \emph{Multi-Loss Rebalancing Algorithm}~\cite{MultiLoss2020} to weight different loss terms automatically and dynamically.

We have applied our \emph{MotionHint} algorithm with multiple SSM-VO methods and evaluated their performance on the standard KITTI~\cite{KITTI2012} dataset.
We improve the performance by reducing the resulting ATE by up to 28.73\%.


\section{Related Work}

\subsection{Supervised VO Methods}

Visual odometry is a classical problem that revolves around estimating ego-motion incrementally with visual inputs. 
Konda et al.~\cite{firstCNN2015} propose the first learning-based method to solve this problem by formulating it as a classification task.
Considering the temporal characteristics in the VO problem, Wang et al.~\cite{DeepVO2017} introduce Recurrent Neural Networks (RNNs) into the architecture and get more accurate results.
Xue et al.~\cite{Xue2018} attempt to learn the process of feature selection and forward the selected features to a dual-branch RNN for final results.


\subsection{Self-Supervised VO Methods}
\label{sec:related workB}

To mitigate the requirement of the ground truth, many self-supervised methods have been proposed.
SfmLearner~\cite{zhou2017} is one of the earliest self-supervised methods, and it predicts depth and ego-motion simultaneously by using view synthesis.
To further obtain the absolute scale, UnDeepVO~\cite{UnDeepVO2018} trains the network with stereo image pairs, but it is only tested with consecutive monocular images.
Supervised by several spatial and temporal constraints, UnDeepVO can generate accurate trajectories with an absolute scale. 
GeoNet~\cite{GeoNet2018} is the first attempt to learn depth, optical flow, and ego-motion using multi-task learning.
To overcome the scale inconsistency of SSM-VO, Bian et al.~\cite{bian2019} propose a geometry consistency loss to constrain the learning process, obtaining more accurate and scale-consistent results.
Recently, inspired by geometric methods, Zou et al.~\cite{zou2020} propose a new self-supervised framework with a two-layer ConvLSTM~\cite{ConvLSTM2015}, which maintains local and global consistency.
Although these methods obtain some positive results, their self-supervised loss function relies on the consistency loss.
The consistency loss refers to a series of losses, which are built by forcing several predicted results to meet certain constraints and all these predicted results can be optimized.
The consistency loss can be low as long as the predicted results meet the constraints even if they are far from the ground truth.
Therefore, existing SSM-VO using consistency losses can easily fall into local minima.

To avoid local minima, DepthHint~\cite{DepthHint2019} takes the depth map into consideration.
The depth map is generated by geometric methods using synchronized stereo images.
This can be costly and inaccurate.
Therefore, DepthHint cannot fully rely on the generated depth map, but adaptively uses the depth map when it can reduce the warping loss.

\begin{figure*}
\includegraphics[scale=0.43]{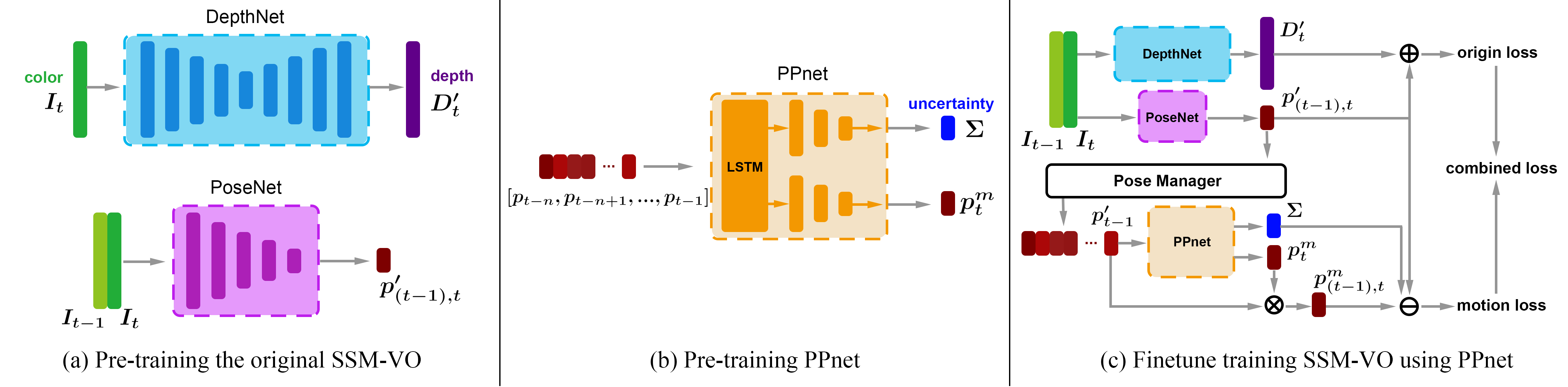}
   \caption{{\bf Overview.} Our \emph{MotionHint} algorithm consists of three training phases. {\bf (a) Pre-training SSM-VO:} We pre-train a SSM-VO and take it as our finetuned object. {\bf (b) Extract the motion model:} \emph{PPnet} is pre-trained to extract the motion model, which can predict the next pose and its uncertainty from a set of consecutive prior poses. {\bf (c) Finetune training the SSM-VO using the motion model:} \emph{PPnet} takes a set of consecutive prior poses saved in \emph{Pose Manager} as input and predicts the pseudo pose. The pseudo pose is further used to build the pseudo label of the current predicted ego-motion. The weighted difference between the pseudo label and the predicted ego-motion generates the motion loss, which guides the SSM-VO out of local minima.
   `$\otimes$' computes the relative pose of two absolute poses; `$\ominus$' computes the weighted difference; `$\oplus$' computes the supervision loss of the original self-supervised system.
    } 
\label{fig:pipeline}
\end{figure*}

\section{Our MotionHint Algorithm}

\subsection{Background}

Supervised monocular VO systems build their loss function using the difference between the ground truth and the predicted ego-motion. 
Without the ground truth, self-supervised methods usually formulate the monocular VO task as a view synthesis problem, by training the networks to synthesize a target image from source images with the estimated depth and ego-motion. Then, the loss function can be built on the difference between the target image and warped images.
However, the loss function built by view synthesis is a kind of consistency loss, which can easily fall into local minima as mentioned in Section \ref{sec:related workB}.

\subsection{Notations}

\begin{itemize}
    \item $I_t$ denotes the input image at time $t$.
    \item $D_t$ denotes the depth map of the image $I_t$.
    \item $p$ denotes a 6-DoF (Degree of Freedom) pose of the camera. $p_t$ denotes the absolute pose of the image $I_t$, and $p_{(t-1),t}$ denotes the relative poses between the image $I_{t-1}$ and $I_t$.
    \item $T$ denotes a transformation matrix of the camera. $T_t$ denotes the absolute transformation matrix of the image $I_t$, and $T_{(t-1),t}$ denotes the relative transformation matrix between the image $I_{t-1}$ and $I_t$.
    \item $(\cdot)^m$ denotes the results generated by our \emph{PPnet}.
    \item $(\cdot)'$ denotes the results predicted by the original SSM-VO.
    \item $L$ denotes the loss function.
\end{itemize}

\subsection{Algorithm Overview} \label{sec:overview}

As shown in Figure \ref{fig:pipeline}, our \emph{MotionHint} algorithm consists of three training phases: (1) pre-training the original SSM-VO, (2) pre-training the \emph{PPnet}, (3) finetune training the pre-trained SSM-VO with the pre-trained \emph{PPnet}.
The first phase can be replaced with existing SSM-VO systems, which is our finetuned object;
the second phase will be described in Section~\ref{sec:ppnet}, which is used to extract the motion model; and the third phase is the key phase to apply the motion model to SSM-VO and will be described in the following paragraph. In the third phase, \emph{PPnet} is weights-fixed.
The inference is just the same as the original SSM-VO without any additional cost.

Figure~\ref{fig:pipeline}(c) shows the data flowchart of the third phase.
During this phase, the original SSM-VO generates the original loss based on predicted depth map $D_t'$ and ego-motion $p_{(t-1),t}'$.
On the other hand, \emph{Pose Manager} gives some of consecutive poses before the target frame to \emph{PPnet}.
Using these poses, \emph{PPnet} predicts the pseudo pose of the target frame $p_t^m$ as well as its uncertainty $\Sigma$ following the learned motion model.
We further generate the pseudo label $p_{(t-1), t}^m$ using the pseudo pose $p_t^m$ and $p_{t-1}'$ in the input sequence.
The motion loss is then computed as the weighted difference between $p_{(t-1),t}'$ and $p_{(t-1),t}^m$.
By combining the motion loss with the original loss (described in Section \ref{sec:MLRA}), we can guide the SSM-VO out of local minima and closer to the global minimum.

\subsection{Pre-training PPnet}\label{sec:ppnet}

\emph{PPnet} is designed to solve the problem of predicting the next pose and its uncertainty using a set of consecutive prior poses.
We formulate the problem as a multivariate time series regression problem: we know a set of consecutive poses $p_1, p_2, ..., p_n$ at an equally-spaced time series $t_1, t_2, ..., t_n$.
The goal is to predict the next pose $p_{n+1}$ and its uncertainty. 

To solve the problem, \emph{PPnet} consists of an LSTM~\cite{LSTM1997} and several linear layers. 
We design our loss function with the methodology of Gast et al. \cite{Lightweight2018} to predict the uncertainty of each dimension.
The overall uncertainty of the predicted pose is the summation of each dimension.

We assume that the outputs of \emph{PPnet} conform to the \emph{general power exponential distribution family}, a multivariate probability density with $d$ dimensions parametrized by three parameters ($\bm{p^m},\bm{\Sigma},k$).
The likelihood function is
\begin{equation}
    p(\bm{y}|\bm{p^m},\bm{\Sigma}) \propto \prod_{j=1}^d \Sigma_j^{-\frac{1}{2}} \exp \left\{ -\frac{1}{2} \left(\sum_{j=1}^d \frac{(y_j - p ^m_j)^2}{\Sigma_j} \right)^k \right\}.
\end{equation}
During training, we minimize the negative log likelihood function with slight changes as
\begin{equation}
    \min L_{prob} = \min \gamma \sum_{j=1}^d \log \Sigma_j + \left( \sum_{j=1}^d \frac{(y_j - p^m_j)^2}{\Sigma_j} \right)^k.
\end{equation}
Here, $p^m_j$ is the element of output values and $\Sigma_j$ is the corresponding uncertainty.
$y_j$ is the element of the ground truth.
$k$ is a hyperparameter of the probabilistic model and is set to $\frac{1}{2}$, as suggested in~\cite{Lightweight2018}.
We add $\gamma$ to reweight the uncertainty regularization term and the residual regression term.
The former prevents the network from predicting too large an uncertainty and the latter provides the real supervision of the regression task~\cite{kendall2017}.
We require that $\gamma$ is less than 1; experiments show that $\gamma=0.1$ is the best choice.

Besides, we highlight some key components of our approach.

\subsubsection{Pose Centralization}

The inputs of \emph{PPnet} in the third phase are predicted poses of the pre-trained SSM-VO.
However, the cumulative error increases with the distance~\cite{VOTutorial2011}, which means that input poses of \emph{PPnet} will be inaccurate.

\begin{figure}
\includegraphics[scale=0.82]{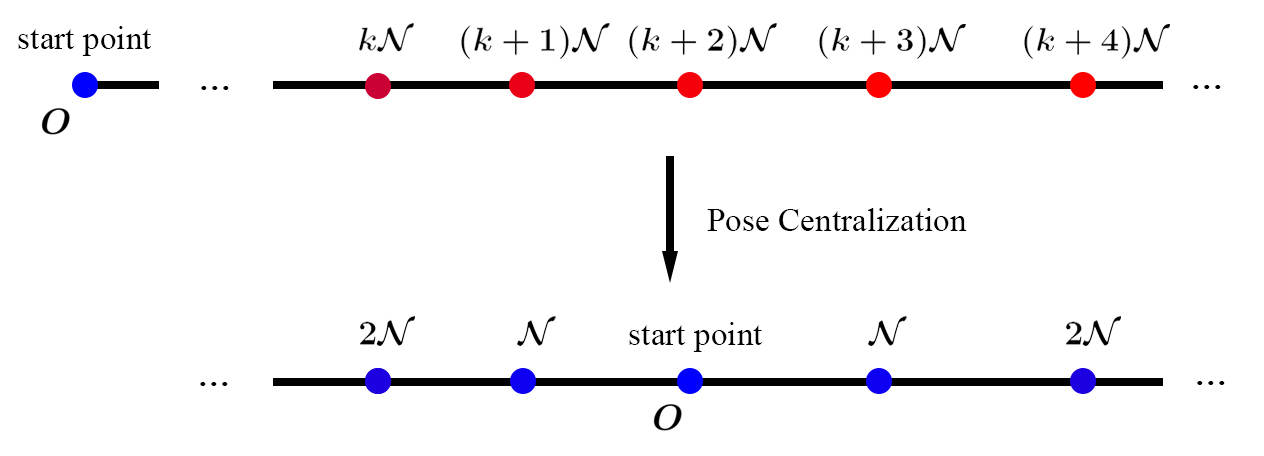}
   \caption{{\bf \emph{Pose centralization}.} $\bm{O}$ refers to the start point where the uncertainty is zero. The intensity of the red color highlights the level of uncertainty. We limit uncertainties of pose sequences to a fixed range by reselecting the starting point.
   } 
\label{fig:pose_centralization}
\end{figure}

To solve this problem, we use \emph{pose centralization} to limit the error of the input sequence to a fixed range by reselecting the starting point of the travel.
Specifically, when inputting a set of poses saved in \emph{Pose Manager} to \emph{PPnet}, we set the central pose of the sequence to a zero vector based on which we recompute other poses and take the processed sequence as new inputs of \emph{PPnet}.
By doing so, during each training step, the traveled distance is limited to a fixed range and so is the error (see Figure~\ref{fig:pose_centralization}).
After \emph{pose centralization}, all central poses of input pose sequences are unified to a zero vector, making the optimization easier.


\subsubsection{Scale Augmentation}


Although \emph{pose centralization} solves the problem of unbounded errors and improves the accuracy of \emph{PPnet}, it cannot handle the issue of scale overfitting. 
During \emph{PPnet} training, if we take original poses as our training dataset without any augmentation, \emph{PPnet} will generate pseudo labels with large uncertainties in the third training phase.

To solve this problem, we employ \emph{scale augmentation}.
For each input pose sequence, we multiply a scale factor to all translation vectors of the sequence.
The scale factor changes for different input pose sequences, and is randomly selected from a manually determined range. 
The selection of the range is important for the performance of the whole algorithm, determining input pose sequences in which scale range \emph{PPnet} can predict well. 
In general, the range is selected to guarantee that the scale of pose sequences used to train \emph{PPnet} covers the scale of poses predicted by SSM-VO.





\subsection{Combining Losses} \label{sec:MLRA}

As shown in Figure ~\ref{fig:pipeline}(c), we have two loss terms in total.
$L_{origin}$ is the original loss of the SSM-VO; $L_{motion}$ is the motion loss built on the weighted difference between the pseudo label and the predicted ego-motion.
The predicted ego-motion $p_{(t-1),t}'$ is the direct output of \emph{PoseCNN} when giving the input frames $I_{t-1}$ and $I_t$ while the pseudo label $p_{(t-1),t}^m$ must be obtained in three steps:
(1) \emph{PPnet} takes a set of consecutive poses saved in the \emph{Pose Manager} before the frame $I_t$ as inputs and predicts the pseudo pose $p_t^m$ as well as its uncertainty.
(2) Using an uncertainty threshold, we determine whether the pseudo pose is confident or not. If not, we skip this training sample.
(3) Using the pseudo pose $p_t^m$ and the predicted pose $p_{t-1}'$ saved in the \emph{Pose Manager}, we can compute the pseudo label as
\begin{equation}
\begin{split}
    & T_t^m = SE(p_t^m),\quad T_{t-1}' = SE(p_{t-1}'), \\
    & p_{(t-1),t}^m = se((T_{t-1}')^{-1} T_t^m);
\end{split}
\end{equation}
$SE(\cdot)$ is used to convert a 6-DoF pose to a transformation matrix and $se(\cdot)$ is the opposite.

Having the pseudo label $p_{(t-1),t}^m$ and the predicted ego-motion $p_{(t-1),t}'$, $L_{motion}$ can be obtained by computing the weighted difference as
\begin{equation}
\label{equ:loss_motion}
    L_{motion} = c||p_{(t-1),t}^m - p_{(t-1),t}'||_2.
\end{equation}
Here, $c$ represents the confidence of the pseudo label, which is a monotonically decreasing function with respect to the uncertainty.

Finally, we define the total loss as
\begin{equation}
\label{equ:combine}
    L = w_1 L_{origin} + w_2 L_{motion}
\end{equation}
and $w_i$ is the weights of different loss terms. 

To determine the weights in a more reasonable way, we employ the \emph{Multi-Loss Rebalancing Algorithm} (MLRA)~\cite{MultiLoss2020}. 
We initialize each weight to $\frac{1}{2}$, and MLRA can update weights periodically during training based on different descending rates of loss terms. 
In this algorithm, a hyperparameter $\lambda$ needs to be specified that indicates the algorithm should focus on the loss term that descends faster or the loss term that descends slower.

However, in order to take the total loss as a criterion for model selection, in our third training phase, we only update the weights once.


\begin{table*}[t]
    \caption{Evaluation results on KITTI of MonoDepth2 and our improved version. Our \emph{MotionHint} algorithm under all setups performs better than baseline and the results under the \emph{Unpaired Pose} setup performs the best.
    ``MH'' indicates whether using \emph{MotionHint}.
    The best performance of each column is in {\bf bold}.
    }
    \centering
    \resizebox{\textwidth}{!}{
    \begin{tabular}{l|c|ccc|ccc}
    \hline
    \multirow{2}{*}{Methods} & \multirow{2}{*}{MH} & \multicolumn{3}{c|}{Seq. 09} & \multicolumn{3}{c}{Seq. 10} \\
     & & $t_{err}(\%)$ & $r_{err}(^\circ/100m)$ & ATE(m) & $t_{err}(\%)$ & $r_{err}(^\circ/100m)$ & ATE(m) \\ \hline
     MonoDepth2 (Baseline) & & 14.837 & 3.299 & 68.180 & 12.097 & 4.949 & 19.408 \\ \hline
     MonoDepth2 + PPnet (Ground Truth) &$\surd$& {13.502} & {2.998} & {62.337} & {10.377} & {4.453} & {17.541} \\
     MonoDepth2 + PPnet (Paired Pose) &$\surd$& {14.071} & {3.099} & {64.704} & {10.976} & {4.495} & {17.752} \\
     MonoDepth2 + PPnet (Unpaired Pose) &$\surd$& \bm{{11.562}} & \bm{{2.601}} & \bm{{54.456}} & \bm{{10.088}} & \bm{{3.949}} & \bm{{15.517}} \\
     \hline
    \end{tabular}
    }
    \label{tab:monodepth2_eval}
\end{table*}

\begin{table*}[t]
    \caption{Evaluation results on KITTI of SC-Depth, our improved version and other SSM-VO systems. 
    ``SC-Depth (Baseline)'' obtains the best performance in open-sourced SSM-VO systems.
    Our \emph{MotionHint} algorithm under all setups performs better than baseline and the results under the \emph{Unpaired Pose} setup even performs better than Zou et al. on sequence 10 about the ATE.
    ``Baseline'' is the latest model of SC-Depth publicly available on github. 
    ``OS'' indicates whether the system is open-sourced. ``MH'' indicates whether using \emph{MotionHint}.
    The best performance of each column is in {\bf bold}, and the second best is \underline{underlined}.
    }
    \centering
    \resizebox{\textwidth}{!}{
    \begin{tabular}{l|c|c|ccc|ccc}
    \hline
    \multirow{2}{*}{Methods} & \multirow{2}{*}{OS} & \multirow{2}{*}{MH} & \multicolumn{3}{c|}{Seq. 09} & \multicolumn{3}{c}{Seq. 10} \\
    & & & $t_{err}(\%)$ & $r_{err}(^\circ/100m)$ & ATE(m) & $t_{err}(\%)$ & $r_{err}(^\circ/100m)$ & ATE(m) \\ \hline
    SfmLearner &$\surd$& & 19.15 & 6.82 & 77.79 & 40.40 & 17.69 & 67.34 \\
    DeepMatchVO &$\surd$& & 9.91 & 3.8 & 27.08 & 12.18 & 5.9 & 24.44 \\
    DW-Learned&$\surd$& & - & - & 20.91 & - & - & 17.88 \\
    DW-Corrected&$\surd$& & - & - & 19.01 & - & - & 14.85 \\
    (Zou et al. 2020) & $\times$ & & \bm{{3.49}} & \bm{{1.00}} & \bm{{11.30}} & \bm{{5.81}} & \bm{{1.8}} & \underline{11.80} \\ \hline
    SC-Depth (Baseline) &$\surd$& & 8.035 & 1.936 & 23.777 & 7.949 & 3.752 & 12.421 \\ \hline
    SC-Depth + PPnet (Ground Truth) &$\surd$&$\surd$& \underline{7.802} & \underline{{1.459}} & \underline{{14.986}} & {8.270} & {2.969} & {11.870} \\
    SC-Depth + PPnet (Paired Pose) &$\surd$&$\surd$& {8.171} & {1.628} & {23.773} & \underline{{7.562}} & {3.170} & {12.055} \\
    SC-Depth + PPnet (Unpaired Pose) &$\surd$&$\surd$& {8.178} & {1.504} & {17.823} & {8.096} & \underline{2.736} & \bm{{11.625}} \\
     \hline
    \end{tabular}
    }
    \label{tab:sc_eval}
\end{table*}

\begin{table*}[t]
    \caption{Ablation study of MonoDepth2. 
    Our \emph{MotionHint} algorithm performs worse than baseline without \emph{PPnet} or any steps of it.
    Our \emph{MotionHint} algorithm can improve the result without MLRA.
    The motion model expressed by \emph{PPnet} has made a major contribution to the improvement.
    \emph{PPnet} here is trained by unpaired poses. 
    The best performance of each column is in {\bf bold}, and the second best is \underline{underlined}.
    `×' indicates the network cannot converge.
    }d
    \centering
    \resizebox{170mm}{!}{
    \begin{tabular}{c|l|ccc|ccc}
    \hline
    \multicolumn{2}{c|}{\multirow{2}{*}{Methods}} & \multicolumn{3}{c|}{Seq. 09} & \multicolumn{3}{c}{Seq. 10} \\
    \multicolumn{2}{l|}{} & $t_{err}(\%)$ & $r_{err}(^\circ/100m)$ & ATE(m) & $t_{err}(\%)$ & $r_{err}(^\circ/100m)$ & ATE(m) \\ \hline
    \multicolumn{2}{l|}{MonoDepth2 (Baseline)} & 14.837 & 3.299 & 68.180 & 12.097 & 4.949 & 19.408 \\ \hline
    \multirow{4}{*}{PPnet} 
    & w/o pose centralization & × & × & × & × & × & × \\
    & w/o scale augmentation & × & × & × & × & × & × \\
    & not using uncertainty & {15.442} & {3.388} & {70.629} & {12.015} & {4.963} & {19.624} \\
    & using uncertainty w/o thres & {15.302} & {3.345} & {69.762} & {11.736} & {4.930} & {19.549} \\ \hline
    \multicolumn{2}{l|}{w/o PPnet (generate pseudo labels using ground truth)} & {15.446} & {3.393} & {70.480} & {12.108} & {4.901} & {19.199} \\
     \hline
     \multicolumn{2}{l|}{w/o \emph{MLRA} (sum two losses up directly) } & \underline{12.966} & \underline{2.882} & \underline{60.490} & \underline{10.890} & \underline{4.365} & \underline{17.104} \\
     \hline
     \multicolumn{2}{l|}{MonoDepth2 + PPnet + MLRA} & \bm{{11.562}} & \bm{{2.601}} & \bm{{54.456}} & \bm{{10.088}} & \bm{{3.949}} & \bm{{15.517}} \\ \hline
    \end{tabular}
    }
    \label{tab:ablation_study}
\end{table*}

\section{Evaluation}

\subsection{Experiment Setup}


We select MonoDepth2~\cite{monodepth2} and SC-Depth~\cite{bian2021} as our baseline SSM-VO systems.
SC-Depth purportedly outperforms all previous monocular alternatives except the most recent approach~\cite{zou2020}.
However, \cite{zou2020} is not open-sourced, so we consider SC-Depth as the best open-sourced SSM-VO system.

We employ the standard KITTI~\cite{KITTI2012} benchmark to evaluate our method.
The KITTI dataset contains 22 sequences captured by the same autonomous driving platform.
Specifically, the ground truth is provided for sequences 00-10, but no ground truth is provided for sequences 11-21.

During the first and third training phases, 00-08 are used for training and 09-10 are used for testing.
This setup is commonly employed by prior SSM-VO systems.
For the second training phases, in which \emph{PPnet} is trained, we prepare 3 different setups to make better ablation comparisons:

\subsubsection{Ground Truth}
Under this setup, we train \emph{PPnet} with the ground truth of sequences 00-08.
This setup requires the ground truth which is difficult to obtain and also the main limitation of supervised methods.

\subsubsection{Paired Pose}
Under this setup, we provide sequences 00, 02-07 to the classic geometric method, monocular ORB-SLAM2~\cite{ORB22017} and train \emph{PPnet}  with the obtained results.
We manually omit sequences 01 and 08 because of their visually inaccurate results.
We use this setup to show that when the ground truth is not available, coarse results generated by geometric methods can be used to train \emph{PPnet}.
But in practice, there is no guarantee that geometric methods can generate accurate results from sequences used for training (e.g., sequence 01 and 08).

\subsubsection{Unpaired Pose}
Under this setup, we provide sequences 11-13, 15-20 to monocular ORB-SLAM2 and train \emph{PPnet} with the obtained results.
We manually omit sequences 14 and 21 because of their visually inaccurate results.
This is the most practical setup since such training sequences are easier to obtain.
Since \emph{PPnet} extracts the motion model, there is no need for it to be trained with the same sequences that trained the original SSM-VO systems.
We use this setup to show that any sequences captured on the same vehicle provide enough information to extract a usable motion model.
Of course, it is only fair that we avoid using sequences 09-10 to train \emph{PPnet}, since they are used for testing.



\subsection{Evaluation Metrics}

For the standard KITTI benchmark, we adopt the average absolute trajectory error (ATE), relative translation errors ($t_{err}\%$), and relative rotation ($r_{err}\%$) errors for all possible subsequences of length (100, 200, ..., 800 meters). Since the absolute scale of the generated trajectory is unknown, we align the trajectory using the evo toolbox~\cite{evo2017}.

\subsection{Parameter Setting}


We employ LSTM~\cite{LSTM1997} and several linear layers to construct our \emph{PPnet} as described in Section~\ref{sec:ppnet}.
Because the inputs of \emph{PPnet} are just a set of 6-DoF poses,
LSTM simply consists of one layer containing 8 hidden units.
Inspired by~\cite{Lightweight2018}, we forward features extracted by LSTM to two-branch linear layers to predict the next pose and its uncertainty.
During each training step, we take 20 consecutive poses as an input sequence of \emph{PPnet}.

We implement our system using the publicly available PyTorch~\cite{pytorch2019} framework.
For all procedures, we train the network with an Adam optimizer~\cite{Adam2014} with $\beta_1 = 0.9, \beta_2 = 0.999$.
The learning rate of \emph{PPnet} is set to 1e-3 while those of the first and third phase are the same as the original settings.
The hyperparameter $\lambda$ of MLRA is set to 1.0.
The period of updating weights in MLRA is 1250 training samples. 

\subsection{Evaluation on MonoDepth2} \label{sec:monodepth2_eval}

To verify that our \emph{MotionHint} algorithm can improve SSM-VO systems in practice, we first perform experiments using the \emph{Unpaired Pose} setup.
The results are shown in the fifth row of Table~\ref{tab:monodepth2_eval}.

To further compare the performance of our MotionHint algorithm with different known information, we conduct experiments using the \emph{Ground Truth} setup and the \emph{Paired Pose} setup.
Results are shown in the third and fourth rows in Table~\ref{tab:monodepth2_eval}.



All results in Table~\ref{tab:monodepth2_eval} show that our \emph{MotionHint} method can greatly improve the performance of MonoDepth2, whether using \emph{Ground Truth}, \emph{Paired Pose}, or \emph{Unpaired Pose}.
The result under the \emph{Unpaired Pose} setup is even better than that under the \emph{Ground Truth} setup, probably because the \emph{Unpaired Pose} setup can provide more training data than the ground truth.
More training data leads to richer motion patterns, and more accurate uncertainty of the predictions, which is exactly the advantage of our \emph{PPnet}.

\subsection{Evaluation on SC-Depth}

To verify the portability of our \emph{MotionHint} algorithm, we also apply our approach on the publicly available SC-Depth~\cite{bian2021}.
We visualize predicted trajectories of the sequence 09 in Figure \ref{fig:teaser}.
Results in Table~\ref{tab:sc_eval} show that SC-Depth obtains the best performance when compared with previous SSM-VO systems, except~\cite{zou2020}.
However, because \cite{zou2020} is not open-sourced, we cannot evaluate our algorithm on their system.
Our \emph{MotionHint} algorithm greatly improves the performance of SC-Depth.
Especially under the \emph{Unpaired Pose} setup, SC-Depth combined with our \emph{MotionHint} algorithm can outperform Zou et al.\cite{zou2020} on sequence 10 about the ATE, although they use a more complex network structure to process temporal information.


\subsection{Ablation Study} \label{sec:ablation_study}

To further demonstrate the role of each component in our method, we perform some ablation studies on MonoDepth2, as shown in Table~\ref{tab:ablation_study}.
\emph{PPnet} in this study is all trained under the \emph{Unpaired Pose} setup.

The first row is our baseline, obtained by continuing to train based on the public model on github with the same learning rate (1e-7) as our third phase.
The following four rows demonstrate some detailed components of \emph{PPnet}.
When trained without \emph{pose centralization}, \emph{PPnet} finds it difficult to extract the motion model and just predicts next poses with large uncertainties. 
Training \emph{PPnet} without \emph{scale augmentation} produces similarly bad results because scale overfitting will also make next poses predicted by \emph{PPnet} have large uncertainties in the third training phase. 
Results in the fourth row show that building the motion supervision without uncertainties will just degrade the performance.
Furthermore, the fifth row shows that if using uncertainties but not using a threshold to filter out pseudo labels with high uncertainties, the self-supervised system can result in worse results than our baseline but better results than not using uncertainties.

In the sixth row, we directly use the difference between the pseudo label generated by the ground truth $p_{(t-1),t}^g$ and the predicted ego-motion $p_{(t-1),t}'$ to build an additional supervision without using \emph{PPnet}.
Here, $p_{(t-1),t}^g$ is obtained by replacing the predicted pose $p_t^m$ with the ground truth in Eqn. 3.
By comparing the results in the sixth row, the last row, and the first row, we find that our \emph{MotionHint} algorithm still gets the best result, while using the ground truth directly will just harm the original SSM-VO systems due to scale inconsistency.
To demonstrate the role of MLRA, we perform an experiment to sum two losses up directly; the results are shown in seventh row.
We find that even without MLRA, 
\emph{PPnet} can still improve the performance, and MLRA is just employed to automatically generate more reasonable weights to make the performance better. 

\begin{figure}
\centering
\includegraphics[scale=0.33]{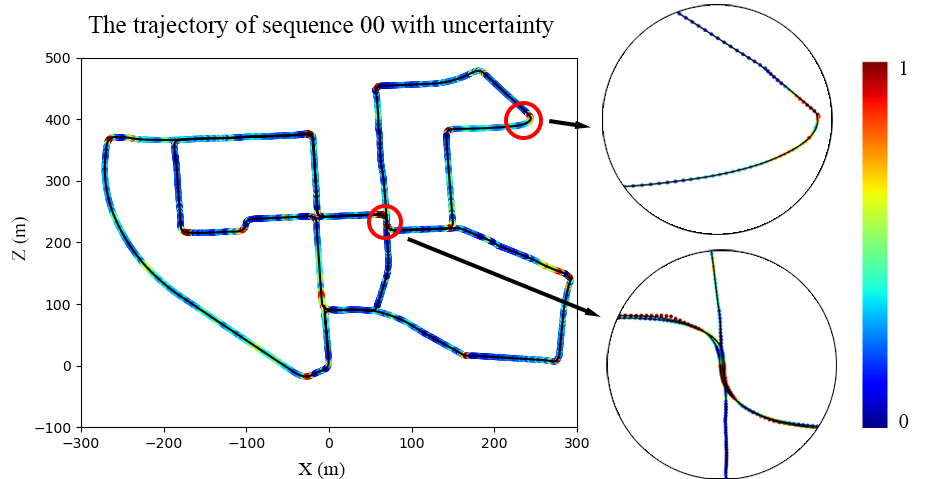}
   \caption{{\bf Qualitative results of \emph{PPnet}.} The black line refers to the ground truth and the colored points refer to poses predicted by \emph{PPnet}. Points in a redder color indicate larger uncertainty.
   } 
\label{fig:ppnet_result}
\end{figure}

\subsection{Effect of PPnet}

We further demonstrate the effect of \emph{PPnet} to show that it can extract a motion model.
In our method, \emph{PPnet} is employed to extract the motion model of the vehicle, predicting the next pose as well as its uncertainty from a set of consecutive prior poses.
Combining proposed \emph{pose centralization} and \emph{scale augmentation}, \emph{PPnet} can predict fairly accurate results.
Qualitative results are shown in Figure~\ref{fig:ppnet_result}.
\emph{PPnet} here is also under the \emph{Unpaired Pose} setup.

We use the ground truth to construct input sequences and predict next poses using \emph{PPnet}.
The trajectory in Figure~\ref{fig:ppnet_result} is generated by these predicted poses.
Here, the black line is the ground truth of sequence 00 and the colored points are poses predicted using \emph{PPnet}. 
Points in a redder color indicate larger uncertainty.
The results show that \emph{PPnet} can predict fairly accurate next poses in most instances except some corners, which we can filter out through a special uncertainty threshold.




\section{Conclusion, Limitation, And Future Work}

In this paper, we have explored the motion model of the vehicle on which a camera is deployed and presented a new self-supervised algorithm, \emph{MotionHint}, to apply this motion model to improve the performance of existing SSM-VO systems.
Our results on the standard KITTI benchmark show that our proposed algorithm can be easily applied to existing SSM-VO systems to significantly improve their performance.
Compared with original SSM-VO systems, our \emph{MotionHint} algorithm can reduce the resulting ATE by up to 28.73\%.

However, the performance of our method depends on the choice of parameters and the quality of generated trajectories by ORB-SLAM2.
In the future, we'd like to attempt the SGP algorithm~\cite{2021SGP}, which performs alternating minimization to train two related networks.
The SGP algorithm can alleviate the limitations of our method and is promising to achieve better results.

{\small
\bibliographystyle{ieee_fullname}
\bibliography{egbib}
}

\end{document}